\documentclass[10pt,twocolumn,letterpaper]{article}

\usepackage{cvpr}              

\usepackage[table]{xcolor}

\usepackage{multirow}

\definecolor{verylightgray}{rgb}{0.88, 0.88, 0.88} 
\definecolor{verylightblue}{rgb}{0.8, 0.9, 1.0} 
\definecolor{lightpurple}{rgb}{0.9, 0.9, 1.0}

\definecolor{cvprblue}{rgb}{0.21,0.49,0.74}
\usepackage[pagebackref,breaklinks,colorlinks,allcolors=cvprblue]{hyperref}

\def\paperID{3652} 
\def\confName{CVPR}
\def\confYear{2026}

\def\tracker{SVLTrack}

\title{Learning to Track Instance from Single Nature Language Description}

\author{Yaozong Zheng\textsuperscript{\rm 1,2}, Bineng Zhong\textsuperscript{\rm 1,2}\thanks{Corresponding author.}, Qihua Liang\textsuperscript{\rm 1,2}$^*$, Shuimu Zeng\textsuperscript{\rm 3}, Haiying Xia\textsuperscript{\rm 1,2}, Shuxiang Song\textsuperscript{\rm 1,2}\\
\textsuperscript{\rm 1}Key Laboratory of Education Blockchain and Intelligent Technology, Ministry of Education\\ Guangxi Normal University, Guilin 541004, China\\
\textsuperscript{\rm 2}University Engineering Research Center of Educational Intelligent Technology\\ Guangxi Normal University, Guilin 541004, China\\
\textsuperscript{\rm 3}University of Southampton, Southampton,  SO17 1BJ, United Kingdom\\
{\tt\small yaozongzheng@stu.gxnu.edu.cn, bnzhong@gxnu.edu.cn,qhliang@gxnu.edu.cn}
}

\begin{document}
\maketitle
\begin{abstract}
How to achieve vision-language (VL) tracking using natural language descriptions from a video sequence \textbf{without relying on any bounding-box ground truth}? In this work, we achieve this goal by tackling \textit{self-supervised VL tracking}, which aims to evaluate tracking capabilities guided by natural language descriptions. We introduce \textbf{\tracker}, a novel self-supervised VL tracker that is capable of tracking any referred object by a language description. Unlike traditional methods that equally fuse all language and visual tokens, we propose an efficient Dynamic Token Aggregation Module, which treats each visual token \textbf{unequally}. The module consists of three main steps: i) Based on an anchor token, it selects multiple important target tokens from the template frame. ii) The selected target tokens are merged according to their attention scores and aggregated into the language tokens, thereby eliminating redundant visual token noise and enhancing semantic alignment. iii) Finally, the fused language tokens serve as guiding signals to extract potential target tokens from the search frame and propagate them to subsequent frames, enhancing temporal prompts and encouraging the tracker to autonomously learn instance tracking from unlabeled videos. This new modeling approach enables the effective self-supervised learning of language-guided tracking representations without the need for large-scale bounding box annotations. Extensive experiments on VL tracking benchmarks show that {\tracker} surpasses SOTA self-supervised methods.

\end{abstract}    
\section{Introduction}
\label{sec:intro}

   \begin{figure}[t]
      \centering
      \includegraphics[width=0.9\columnwidth]{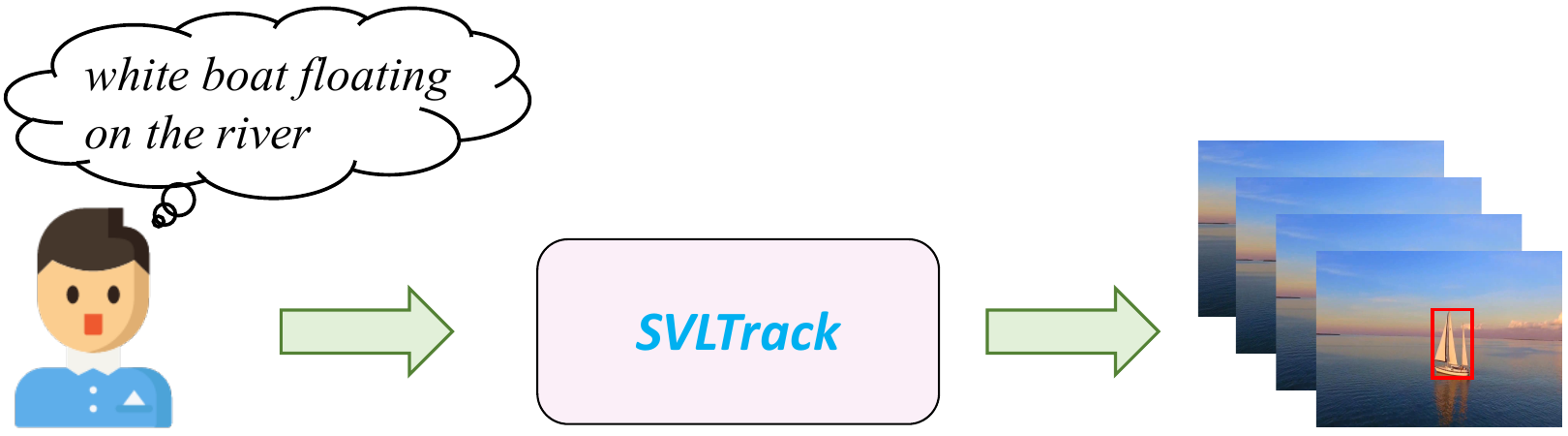}
       \caption{Natural language descriptions can refer to any visual object within a video sequence and continuously track it. Inspired by this idea, we propose {\tracker}, which relies solely on a natural language description for object tracking.
       }
       \label{fig:motivation}
    \end{figure}

Visual object tracking relies on explicit visual prompt, such as initial bounding box, to identify and track target objects in video sequences. Compared to specifying object bounding boxes, natural language descriptions provide a more intuitive and cost-effective way to express human intent in the real world \cite{trackNL}. Therefore, exploring language-guided visual tracking presents an opportunity to enhance the human-computer interaction capabilities of next-generation vision-language (VL) tracking models.

\begin{table}[t]
\centering
\resizebox{\linewidth}{!}{
\begin{tabular}{l|cc|cc}
\toprule
\multicolumn{1}{c|}{\multirow{2}{*}{Tracking Tasks}} & \multicolumn{2}{c|}{Train Annotations} & \multicolumn{2}{c}{Test Initialization} \\
& Multi-Boxes & Language & Init-Box & Language \\
\midrule
Fully-supervised V & \checkmark &  & \checkmark &  \\
Fully-supervised VL & \checkmark & \checkmark & \checkmark & \checkmark \\
\rowcolor{blue!10}
\textbf{Self-supervised VL} & & \checkmark &  & \checkmark \\
\bottomrule
\end{tabular} }
\caption{Comparison with different tracking tasks. \textit{V} stands for the vision-only tracking task, while \textit{VL} represents the vision-language tracking task.}
\label{tab:tasks}
\end{table}

To achieve this challenging goal, recent studies \cite{jointNLT,UVLTrack,mmtrack,queryNLT} have developed robust VL tracking models by introducing natural language descriptions. Typically, these methods fine-tune the multi-modal fusion operations under the supervision of bounding box annotations to ensure stable multi-modal tracking performance. For example, the fully supervised JointNLT \cite{jointNLT} introduces a multi-source relation modeling module and performs vision-language fusion and training on the LaSOT \cite{lasot}, TNL2K \cite{tnl2k}, and OTB99 \cite{trackNL} datasets, which contain 3.52M, 1.24M, and 59K bounding box annotations, respectively, achieving AUC scores of 56.9\%, 54.6\%, and 59.2\%. Although this fine-tuning training strategy appears effective, it obscures the following drawbacks: (1) These fully supervised VL tracking methods heavily relies on thousands of bounding box annotations, which incur substantial time and labor costs. (2) They \textit{equally} utilize the full set of visual and linguistic tokens for vision-language fusion, which leads to redundant computations and hinders effective alignment between visual and linguistic representations.

\textit{Compared with frame-by-frame bounding box annotation, providing a natural language description for each video sequence is far more efficient and considerably reduces human efforts}, as shown in Tab.\ref{tab:tasks}. Thus, in this work, we address the above questions by tackling self-supervised vision-language (VL) tracking, which aims to accurately predict the target localization for each video frame based on a \textbf{\textit{single}} implicit language description. 
By exploring a new VL tracking framework under a language-guided self-supervised learning paradigm, we attempt to quantify the semantic guidance contribution of natural language to the tracking model.
Meanwhile, this enables our VL tracking algorithm to interact effectively with objects in dynamic environments based on user instructions, eliminating the reliance on manually bounding boxes.

Furthermore, we propose a novel self-supervised visual tracking framework, named \textbf{\tracker}, which explores the dynamic aggregation of language and visual tokens to improve the self-supervised VL tracking performance, as illustrated in Fig.\ref{fig:motivation}.
In the context of unlabeled video data, directly constructing a VL tracking model using only natural language and video frames is highly challenging. 
Therefore, we leverage large vision-language model (LVLM) enriched with extensive world knowledge, to generate pseudo-bounding boxes for unlabeled videos.
Then, unlike traditional methods that equally fuse all language and visual tokens, we propose an efficient Dynamic Token Aggregation Module, which treats each visual token \textbf{unequally}. The module consists of three main steps:
i) Based on an anchor token, it selects multiple important target tokens from the template frame.
ii) The selected target tokens are merged according to their attention scores and aggregated into the language tokens, thereby eliminating redundant visual token noise and enhancing semantic alignment.
iii) Finally, the fused language tokens serve as guiding signals to extract potential target tokens from the search frame and propagate them to subsequent frames, enhancing temporal prompts and encouraging the tracker to autonomously learn instance tracking from unlabeled videos.
This novel modeling paradigm enables the {\tracker} to effectively acquire language-guided visual tracking representations in a self-supervised fashion, thereby enhancing its robustness and discriminative capabilities in scenarios with weak target cues.
Extensive experiments conducted on VL tracking benchmarks demonstrate that our approach achieves outstanding performance. 

In summary, our contributions are as follows:

\begin{itemize}
\item We introduce a novel self-supervised VL tracking algorithm, called {\tracker}, which aims to explore the capability of natural language to guide visual tracking in the absence of bounding box annotations.

\item We design an efficient dynamic token aggregation module, which aggregates selected key visual tokens with linguistic features, enabling the model to learn VL instance tracking capabilities from unlabeled videos. 

\item Our tracker achieves outstanding performance on four VL tracking benchmarks, including LaSOT, TNL2K, LaSOT$_{\rm{ext}}$, and OTB99. 
\end{itemize}

   \begin{figure}[t]
      \centering
      \includegraphics[width=0.9\linewidth]{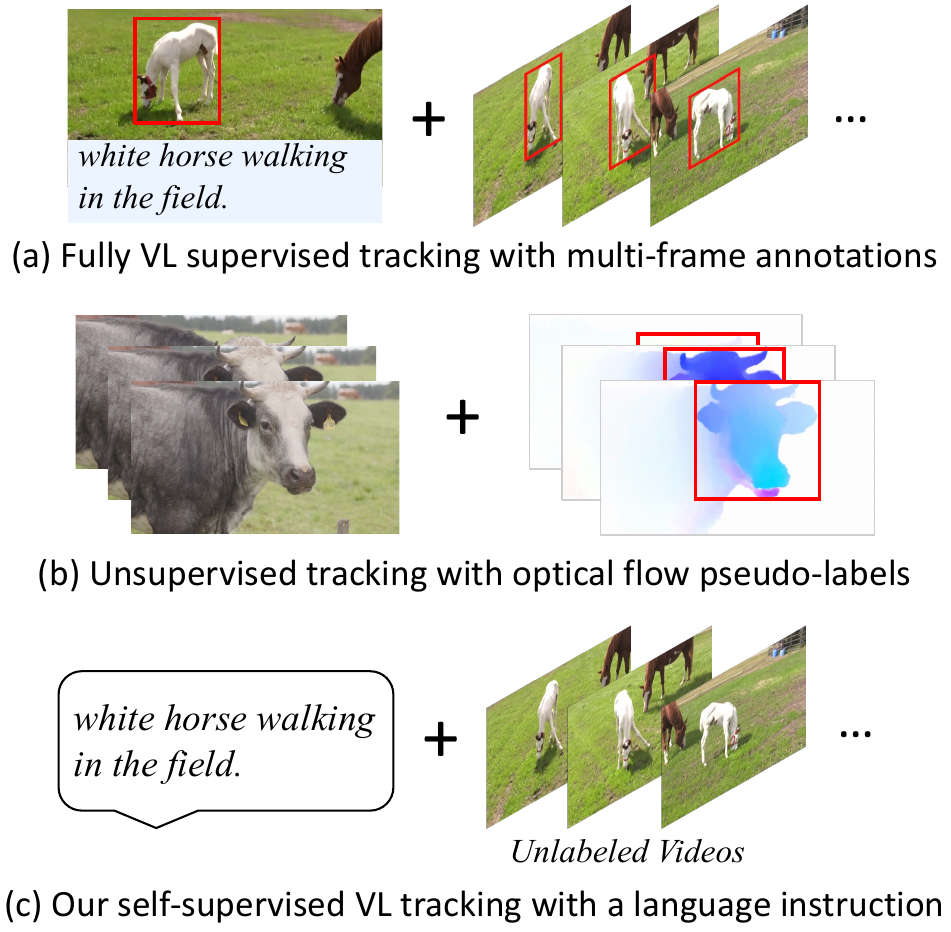}
       \caption{The annotation requirements for different tracking tasks.}
       \label{fig:annos}
    \end{figure}

\section{Related Work}
\label{sec:works}

\subsection{Fully-Supervised Vision-Language Tracking}
Building upon visual tracking \cite{stark,mixformer,UMODTrack,SiamPIN}, vision-language (VL) tracking \cite{trackNL} aims to incorporate natural language descriptions to enhance the semantic awareness of tracking model for target object, such as object categories and attribute information. 
Wang \cite{tnl2k} introduce a large-scale benchmark dataset called TNL2K, providing a foundational dataset for research in the VL tracking community. 
UVLTrack \cite{UVLTrack} proposes a modality-unified feature extractor to jointly train visual and vision-language tracking tasks.
DTLLM \cite{DTLLM} leverages large language models to generate new natural language descriptions for VL tracking datasets, providing multi-granularity semantic information.
chatTracker \cite{chattracker} uses the multimodal large language model to generate high-quality language descriptions and enhance tracking performance.
DUTrack \cite{DUTrack} leverages a large language model to generate dynamic language descriptions for the object based on visual features.

Despite the significant success of the above methods, they heavily rely on large-scale labeled datasets for training. To fully leverage the abundance of unlabeled video data available on the internet, there is a growing interest in exploring self-supervised and unsupervised tracking. These methods hold the potential to reduce the reliance of tracking algorithms on annotated bounding boxes.

\subsection{Self-supervised and Unsupervised Tracking}
Unlike fully supervised tracking, self-supervised and unsupervised tracking tasks aim to develop general object tracking algorithms with minimal or no target supervision signals, thereby reducing annotation costs. Wang \cite{UDT} introduces the cycle consistency with a consistency loss to the tracking domain and propose the first unsupervised discriminative tracker. USOT \cite{USOT} utilizes an optical flow model \cite{optical} to generate pseudo-labels for tracking datasets and employs a multi-stage training process to develop an unsupervised Siamese tracker.  
ULAST \cite{ULAST} introduces a differentiable region mask to select accurate target features, ensuring training stability. Diff-Tracker \cite{DiffTracker}, based on a pre-trained diffusion model \cite{diffusionModel}, designs an initial prompt learner to learn target prompts, enabling high-performance unsupervised tracking.

On the other hand, exploration in self-supervised tracking remains relatively limited. Self-SDCT \cite{SDCT} and TADS \cite{TADS} build self-supervised tracking frameworks using cycle consistency and contrastive learning techniques, respectively. SSTrack \cite{sstrack} decouples forward and backward tracking into global and local tracking, respectively, enabling separate processing of unlabeled and labeled video frames to enhance performance.
Ge \cite{ATTracker} fine-tune large language model \cite{x2VLM} using initial information (natural language descriptions and bounding boxes) to generate redundant pseudo-labels. They then employ a multi-stage training strategy to develop a semi-supervised VL tracking model.
However, its pseudo-label generation stage requires fine-tuning and involves a relatively complex multi-stage training strategy, making it susceptible to noisy pseudo labels and resulting in limited performance gains.
In contrast to these efforts, we solely utilize natural language descriptions as target reference information and propose a new self-supervised VL tracking framework to autonomously learn instance tracking from unlabeled videos.

\subsection{Large Vision-Language Models}

In recent years, driven by the excellent reasoning capabilities of large vision-language models (LVLMs), an increasing number of researchers explore extending these capabilities to various visual domains. 
For instance, 
BLIP-2 \cite{BLIP2} employs a Q-Former to project features from a visual encoder for integration into a large language model (LLM). 
LISA \cite{LISA} combines a large vision-language model with a segmentation decoder to enable flexible reasoning segmentation. Inspired by LISA, VISA \cite{VISA} and TrackGPT \cite{TrackGPT} extend its concepts to the fields of video object segmentation and tracking, respectively. 
APE \cite{APE} leverages the aligning and prompting paradigm to construct an instance-level vision-language foundation model.
In contrast, our work focuses on incorporating the reasoning capabilities of large vision-language models into the VL tracking field, achieving more user-friendly and open vision-language tracking.

   \begin{figure}
      \centering
      \includegraphics[width=0.9\linewidth]{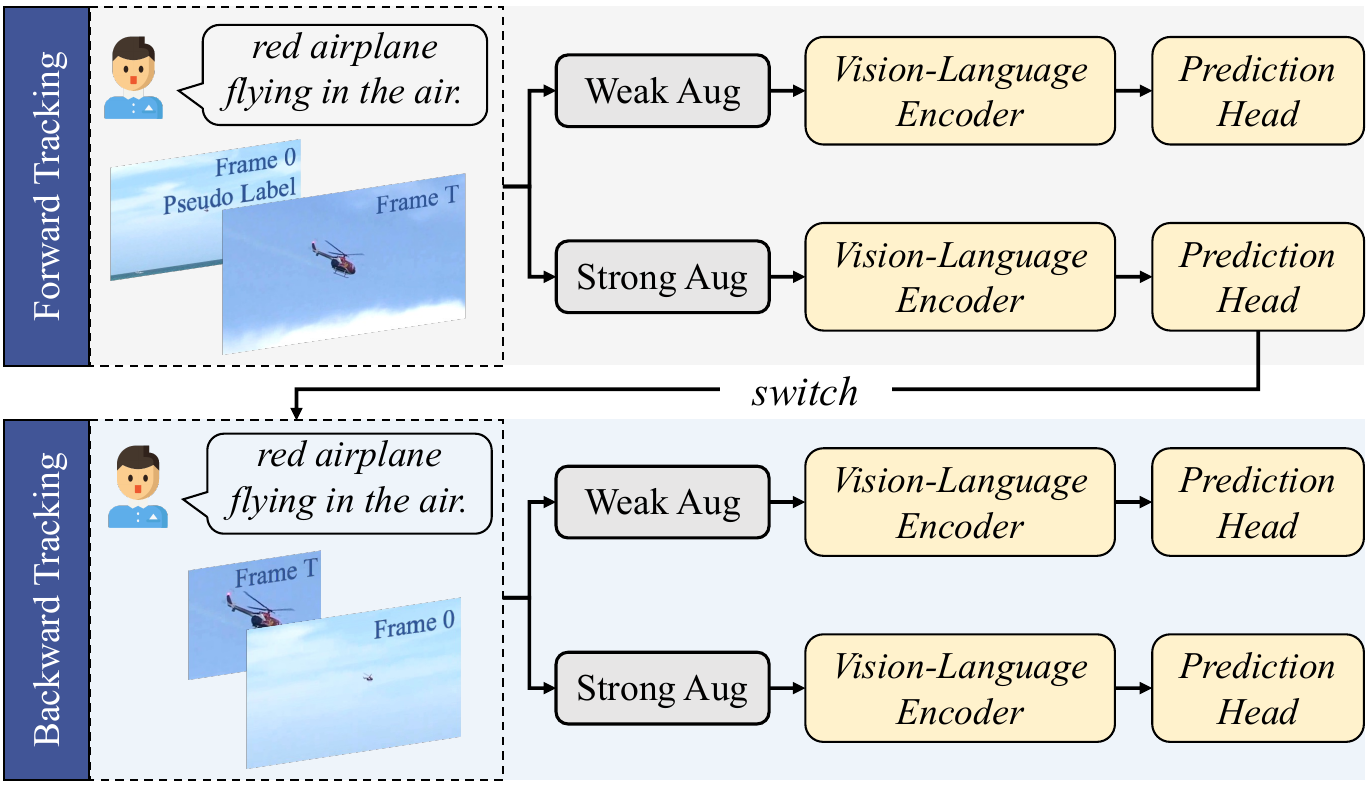}
       \caption{Overview of the proposed {\tracker} framework.}
       \label{fig:framework}
    \end{figure}

   \begin{figure*}
      \centering
      \includegraphics[width=0.9\linewidth]{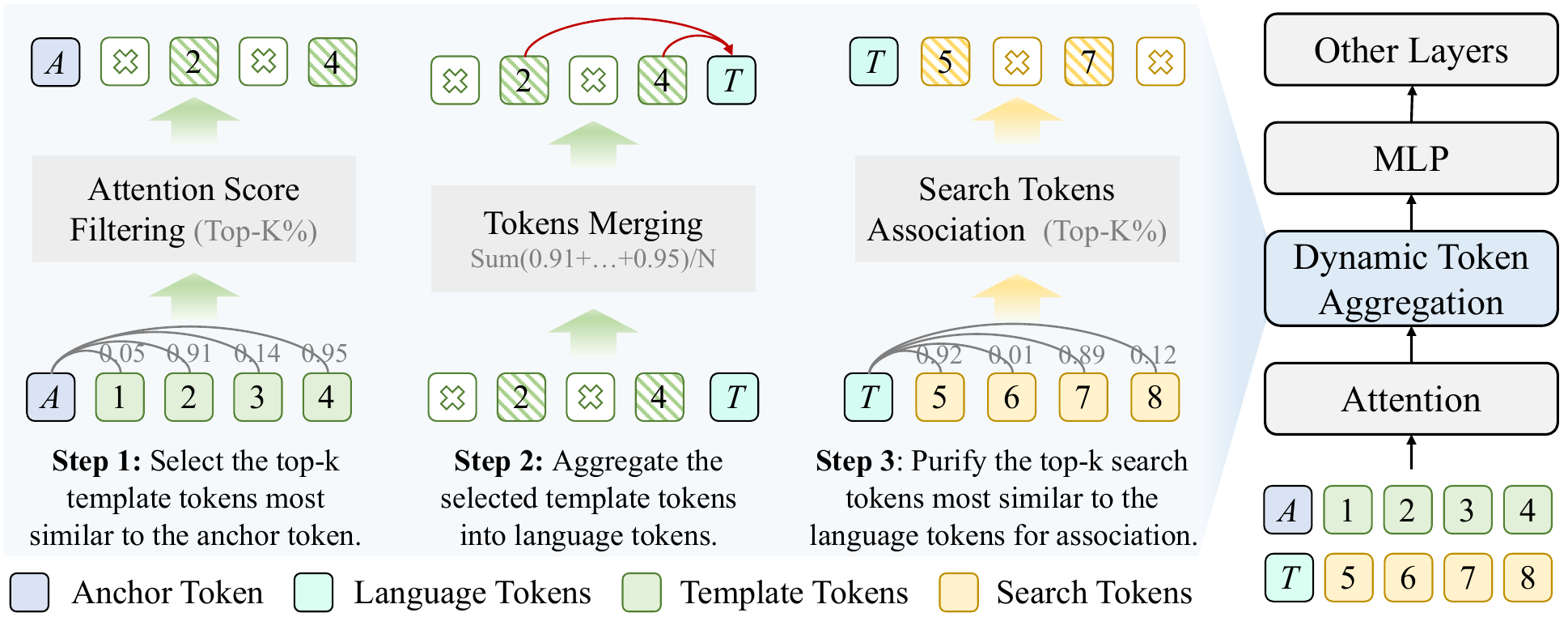}
       \caption{
       \textit{Our Dynamic Token Aggregation Module}.
       It treats each visual token unequally, following three main steps:
       i) Based on an anchor token, it selects multiple important target tokens from the template frame.
       ii) The selected target tokens are merged according to their attention scores and aggregated into the language tokens.
       iii) The fused language tokens serve as guiding signals to extract potential target tokens from the search frame and propagate them to subsequent frames, enhancing temporal prompts and encouraging the tracker to autonomously learn instance tracking from unlabeled videos.
       }
       \label{fig:merging}
    \end{figure*}

\section{Methodology}
\subsection{Task Definition}

Li \etal \cite{trackNL} first propose \textbf{fully supervised} VL tracking task and introduce three VL tracking configurations: 1) one relying on lingual target specification only, 2) one relying on visual target specification based on language, and 3) one leveraging their joint capacity. 
In this work, as shown in Fig.\ref{fig:annos}, we extend the second tracking configuration above into a language-guided \textbf{self-supervised} VL tracking by discarding all bounding box annotations.
This task integrates language reasoning with long-term video understanding to localize the target object in videos accurately. 
Specifically, given a natural language description, the self-supervised VL tracking first identifies the referred initial target object and then continuously tracks it across subsequent frames.
This tracking paradigm is designed to rely on minimal linguistic cues, establishing a cost-effective and user-friendly VL tracking model. 
Specifically, given a natural language instruction $Q_{\text{text}}$ describing an arbitrary instance, and an input video clip $\mathcal{V} \in \mathbb{R}^{T \times H \times W \times 3}$, the task aims to train a tracking model $\varphi$ from \textit{unlabeled videos}, enabling it to accurately locate the target in subsequent frames. Formally, the task can be defined as:
   \begin{equation}
      \mathcal{B} = \varphi \left(\mathcal{V}, Q_{\text{text}} \right),
      \label{eq:self_supervised}
   \end{equation}
where the input video clip $\mathcal{V} = \{I\}_{t=1}^T$ contains $T$ frames. The output target bounding boxes $\mathcal{B} = \{B\}_{t=1}^T$ has the same frame number.
Notably, for the self-supervised tracking model, the target cues consists solely of a single language instruction, without referencing the initial frame, and annotations for all video frames remain unseen. This setup necessitates that the tracker demonstrates strong cross-modal understanding and autonomous tracking capabilities.

\subsection{Weak-to-Strong Consistency Self-supervised VL Tracking Pipeline}

Since bounding box annotations are unavailable for self-supervised tracking model, we cannot train model in the same manner as fully-supervised VL tracking algorithms. 
To address this challenge, we leverage large vision-language model (LVLM, i.e., APE \cite{APE} or LISA \cite{LISA}) enriched with extensive world knowledge, to generate pseudo-bounding boxes for the initial frame $I_{0}$ in unlabeled videos based on natural language instruction $Q_{\text{text}}$.
The generation process is formulated as follows:
   \begin{equation}
      \mathcal{B}_{0} = LVLM \left(I_{0}, Q_{\text{text}} \right),
      \label{eq:gen_box}
   \end{equation}
where $\mathcal{B}_{0}$ denotes the generated initial pseudo bounding box.
It should be noted that the language descriptions provided in current VL tracking benchmarks are mostly semantically aligned with the initial frame. However, as time passes, these language descriptions gradually fail to match the appearance of the target or changes in motion. Therefore, we generate pseudo label only for the initial frame.

Furthermore, following the principle of temporal cycle consistency, we design a new weak-to-strong consistency self-supervised VL tracking pipeline, termed {\tracker}, which incorporates both forward and backward tracking, as shown in Fig.\ref{fig:framework}. 
Specifically, each unlabeled frame $I$ is augmented in two ways: a weak augmentation $A^w$ such as center jitter, and a strong augmentation $A^s$ such as color jitter.
We aim to ensure that the predictions from the strongly augmented frame remain as consistent as possible with those from the weakly augmented frame, making full use of the unlabeled video data.
The {\tracker} comprises four main components: a visual encoder $\mathcal{E}_{v}$, a language encoder $\mathcal{E}_{l}$, a dynamic token aggregation module $\mathcal{M}$, and a prediction head $\mathcal{H}$.
Specifically, the framework takes as input a natural language description $Q_{\text{text}}$, a template frame $Z$ with pseudo bounding box, and a search clip $S$. These inputs are first processed through a tokenizer or an embedding layer to generate linguistic embeddings $ \mathcal{F}_{l} \in \mathbb{R}^{N_{l} \times D}$, template features $ \mathcal{F}_{z} \in \mathbb{R}^{N_{z} \times D}$ and search features $ \mathcal{F}_{s} \in \mathbb{R}^{N_{s} \times D} $, which are then fed into multiple transformer layers for feature extraction and fusion.
In this process, we define an anchor token initialized to zero to learn the appearance representation of the target instance.
Subsequently, the anchor token, linguistic embeddings and visual features are passed into the dynamic token aggregation module to analyze the importance of visual tokens and enable accurate multi-modal integration. Finally, the fused visual representations are fed into the prediction head to perform target localization and tracking. Through the above operations, we can perform a complete forward tracking process, which can be formulated as follows:
   \begin{equation}
        \begin{split}
        \mathcal{F}_{vl} &= \mathcal{M} \left( \mathcal{E}_{v}\left(\mathcal{V} \right), \mathcal{E}_{l} \left( Q_{\text{text}} \right) \right), \\
        \mathcal{B} &= \mathcal{H} \left( \mathcal{F}_{vl} + \mathcal{E}_{v}\left(\mathcal{V} \right) \right).
        \end{split}
        \label{eq:forward_tracking}
    \end{equation}

Moreover, the backward tracking process is achieved by interchanging the roles of the template and search frames, followed by re-executing the entire network pipeline.
The overall objective function is a combination of supervised loss $\mathcal{L}_{s}$ and unsupervised loss $\mathcal{L}_{u}$ as:
   \begin{equation}
        \begin{split}
        \mathcal{L}_{total} &= \mathcal{L}_{s} + \mathcal{L}_{u},
        \end{split}
        \label{eq:total_loss}
    \end{equation}
where $\mathcal{L}_{s}$ is computed only on the initial frame, while $\mathcal{L}_{u}$ is calculated on the unlabeled video frames.

\subsection{Dynamic Token Aggregation Module}
\label{sec:framework}

The previous multi-modal fusion of visual and linguistic features introduces redundant computation since not all visual tokens are attended to by the multi-head self-attention (MHSA) mechanism. In this section, as shown in Fig.\ref{fig:merging}, we propose an efficient dynamic token aggregation module (DTA) to alleviate this problem, aiming to dynamically identify the most informative visual target tokens and facilitate semantic alignment between visual and language representations. Specifically, our DTA is positioned between the MHSA layer and the multi-layer perceptron (MLP) layer. It operates through three key steps to selectively aggregate the most discriminative visual tokens.

\textbf{STEP 1: Select the important target tokens.} We first concatenate the anchor, language, template, and search tokens along the spatial dimension and feed them into a multi-head attention layer to compute their attention score matrix. The anchor token, represented as a one-dimensional vector, serves as a reference to measure the relative importance of all visual and linguistic tokens. Specifically, we calculate the attention scores between the anchor token and all template tokens to assess the significance of each template token. The tokens with the highest scores are then selected as the most informative target tokens, while those with lower scores, which indicate weaker semantic alignment with the natural language description, are filtered out. The selection process is formally defined as follows:
   \begin{equation}
        \begin{split}
        T_{z} &= TopK \left( \mathcal{F}_{z}, Attn_{az} \right),
        \end{split}
        \label{eq:step1}
    \end{equation}
where $Attn_{az}$ denotes the cross-attention matrix between the anchor token and the template frame, while $T_{z}$ represents the selected target tokens from the template frame.
To balance the modal tokens, the number of visual tokens $T_{z}$ is dynamically adjusted to match that of the language tokens.

\textbf{STEP 2: Aggregate target tokens into language tokens.}
We aim to achieve an efficient and compact fusion of the most valuable visual tokens with linguistic representations. Specifically, the target tokens selected in the previous stage are first merged according to their attention scores to preserve the most discriminative visual target information. The merged visual tokens are then further aggregated into the language tokens, thereby establishing a tighter semantic connection between the visual and linguistic modalities. Through the above token merging strategy, our model reduces redundant multi-modal fusion computations and strengthens the alignment between linguistic descriptions and visual targets. The formula is as follows:
   \begin{equation}
        \begin{split}
        \mathcal{F}_{vl} &= Merging \left( T_{z}, \mathcal{F}_{l} \right),
        \end{split}
        \label{eq:step2}
    \end{equation}
where $\mathcal{F}_{vl}$ denotes the fused vision-language features.

\textbf{STEP 3: Purify the search tokens for temporal association.}
The language tokens fused in the previous stage serve as guiding signals to identify potential target tokens within the search frame. By leveraging the semantic information encoded in these language tokens, the model can effectively filter out irrelevant or noisy visual tokens, selecting those most likely to correspond to the target object. The purified target tokens are then propagated to subsequent frames, enhancing temporal target prompts. By integrating this temporal association mechanism, our self-supervised tracking model is able to autonomously learn instance-level tracking directly from unlabeled video data, maintaining strong robustness even in challenging scenarios.
The formula is described as follows:
   \begin{equation}
        \begin{split}
        T_{s} &= TopK \left( \mathcal{F}_{s}, Attn_{ls} \right),
        \end{split}
        \label{eq:step3}
    \end{equation}
where $Attn_{ls}$ denotes the cross-attention matrix between the lanuage token and the search frame. $T_{s}$ represents the purified target tokens from the search frame.
This novel modeling paradigm enables {\tracker} to effectively acquire language-guided tracking representations in a self-supervised fashion, thus enhancing its robustness and discriminative capabilities in scenarios with weak target cues.

   \begin{table*}[t]
    \centering
    \caption{Comparison with state-of-the-art methods on three popular benchmarks.
    \textit{Init.Info} refers to using a combination of natural language descriptions and an initial bounding box as supervision signals. \textit{Init.Lang} and \textit{Init.BBox} represent using only natural language descriptions or bounding boxes, respectively, as supervision signals.
    Best in \textbf{bold}, second best \underline{underlined}.}
    \resizebox{\textwidth}{!}{
    \begin{tabular}{c|l|c|c|c|ccc|ccc|cc}
    \toprule
    \multicolumn{1}{c|}{\multirow{2}{*}{Type}} & \multicolumn{1}{c|}{\multirow{2}{*}{Method}} & \multicolumn{1}{c|}{\multirow{2}{*}{Source}} & \multicolumn{1}{c|}{\multirow{2}{*}{Supervised}} & \multicolumn{1}{c|}{\multirow{2}{*}{Initialize}}
      & \multicolumn{3}{c|}{TNL2K} &\multicolumn{3}{c|}{LaSOT} & \multicolumn{2}{c}{OTB99} \\ \cline{6-13}
      & & & & & AUC & P${_{\rm{Norm}}}$ & P & AUC & P${_{\rm{Norm}}}$ & P & AUC & P \\
      \midrule
      \multicolumn{1}{c|}{\multirow{15}{*}{\rotatebox{90}{Vision-Language}}} 
      & SNLT \cite{SNLT} & CVPR2021 & Yes & NL+BBox & - & - & - & 54.0 & 63.6 & 57.4 & 66.6 & 84.8 \\ 
      & TNL2K-II \cite{tnl2k} & CVPR2021 & Yes & NL+BBox & 42.0 & 50.0 & 42.0 & 51.3 & - & 55.4 & 68.0 & 88.0 \\ 
      & VLT$_{\rm{TT}}$ \cite{VLT} & NeurIPS2022 & Yes & NL+BBox & 54.7 & 71.8  & 55.3 & {67.3} & {80.2} & {71.5} & {74.0} & 89.8 \\ 
      & MMTrack \cite{mmtrack} & TCSVT2023 & Yes & NL+BBox & 58.6 & 75.2 & 59.4 & 70.0 & {82.3} &{75.7} & {70.5} & {91.8} \\
      & JointNLT \cite{jointNLT} & CVPR2023 & Yes & NL+BBox & {56.9} & {73.6} & {58.1} & 60.4 & 69.4 & 63.6 & 65.3 & 85.6 \\
      & UVLTrack \cite{UVLTrack} & AAAI2024 & Yes & NL+BBox & 63.1 & - & 66.7 & 69.4 & - & 75.9 & 69.3 & 89.9 \\
      & ATTracker \cite{ATTracker} & ACM MM2024 & Yes & NL+BBox & 56.9 & 75.0 & 64.7 & 63.7 & - & 67.3 & 69.3 & 90.3 \\
      & DTLLM \cite{DTLLM} & CVPRW2024 & Yes & NL+BBox & 56.6 & - & 56.9 & 69.0 & - & 74.7 & 70.6 & 91.1 \\
      & MemVLT \cite{MemVLT} & NeurIPS2024 & Yes & NL+BBox & 63.3 & 80.9 & 67.4 & 72.9 & 85.7 & 80.5 & - & - \\
      & JointNLT \cite{jointNLT} & CVPR2023 & Yes & NL & 54.6 & - & 55.0 & 56.9 & - & 59.3 & 59.2 & 77.6 \\
      & UVLTrack \cite{UVLTrack} & AAAI2024 & Yes & NL & 55.7 & - & 57.2 & 57.2 & - & 61.0 & 60.1 & 79.1 \\
      & ATTracker \cite{ATTracker} & ACM MM2024 & \textit{Init.Info} & NL+BBox & 40.6 & 56.7 & 40.9 & 48.9 & - & 47.5 & 55.5 & 77.5 \\ 
      & DUTrack \cite{DUTrack} & CVPR2025 & Yes & NL+BBox & 64.9 & 82.9 & 70.6 & 73.0 & 83.8 & 81.1 & 70.9 & 93.9 \\ 
      \rowcolor{blue!10}
      & \textbf{{\tracker}-L256} & Ours & \textit{Init.Lang} & NL & \underline{42.5} & \underline{58.1} & \underline{42.0} & \underline{52.5} & \underline{60.0} & \underline{52.3} & \underline{65.4} & \underline{87.7} \\ 
      \rowcolor{blue!10}
      & \textbf{{\tracker}-L384} & Ours & \textit{Init.Lang} & NL & \textbf{43.9} & \textbf{58.6} & \textbf{44.2} & \textbf{53.9} & \textbf{61.2} & \textbf{53.9} & \textbf{66.7} & \textbf{89.3} \\ 
      \midrule
      \multicolumn{1}{c|}{\multirow{10}{*}{\rotatebox{90}{Vision-only}}} 
      & ECO \cite{ECO} & CVPR2017 & No & BBox & - & - & - & 32.4 & - & 30.1 & - & - \\
      & UDT \cite{UDT} & CVPR2018 & No & BBox & 27.0 & 37.0 & 31.4 & - & - & - & 59.4 & 30.1 \\
      & LUDT \cite{LUDT} & IJCV2021 & No & BBox & - & - & - & 26.2 & - & 23.4 & 60.2 & 76.9 \\
      & USOT \cite{USOT} & ICCV2021 & No & BBox & 30.0 & 44.1 & 35.7 & 33.7 & - & 32.5 & 58.9 & 80.6 \\
      & ULAST \cite{ULAST} & CVPR2022 & No & BBox & - & - & - & 47.1 & - & 45.1 & 64.8 & 87.9 \\
      & TADS \cite{TADS} & TNNLS2023 & \textit{Init.BBox} & BBox & - & - & - & 45.5 & 54.2 & 44.8 & 65.3 & - \\
      & Diff-Tracker \cite{DiffTracker} & ECCV2024 & No & BBox & - & - & - & 48.6 & - & 47.2 & 66.1 & 89.8 \\
      & SSTrack-256 \cite{sstrack} & AAAI2025 & \textit{Init.BBox} & BBox & 52.1 & - & 53.3 & 64.8 & 75.2 & 69.2 & 67.9 & - \\
      \rowcolor{blue!10}
      & \textbf{\tracker-V256} & Ours & \textit{Init.BBox} & BBox & \underline{52.6} & \underline{70.7} & \underline{54.1} & \underline{65.1} & \underline{76.1} & \underline{69.8} & \underline{67.9} & \underline{91.2} \\
      \rowcolor{blue!10}
      & \textbf{\tracker-V384} & Ours & \textit{Init.BBox} & BBox & \textbf{55.5} & \textbf{73.1} & \textbf{57.8} & \textbf{66.8} & \textbf{77.3} & \textbf{72.2} & \textbf{70.6} & \textbf{93.2} \\  
    \bottomrule
    \end{tabular}
    }
    \label{tab:results}
\end{table*}

\subsection{Denoising Training Strategy}

Given that the generated pseudo-labels inevitably contain coarse or inaccurate object bounding boxes, we further propose a denoising training strategy to improve the stability of self-supervised learning.
Specifically, similar to transformer trackers \cite{ostrack,odtrack}, our {\tracker} produces a classification score map $\mathcal{P}_c$ and a regression score map $\mathcal{P}_r$ for each frame during training.
We compute the euclidean distance $\mathcal{D}$ between the classification score map of the strongly augmented frame and the pseudo gaussian map $\mathcal{G}$ generated from the predicted bounding box of the weakly augmented frame, and use this distance to rank the results accordingly.
The top K\% of samples with the largest distances, identified as noisy samples, are discarded from the loss computation. 
Formally, the formula is as follows:
   \begin{equation}
      \mathcal{D}(\mathcal{P}_c, \mathcal{G}) = \left \| \mathcal{P}_c - \mathcal{G} \right \|_2^2.
      \label{eq:l2}
   \end{equation}

For normal training samples, we jointly optimize the classification and regression score maps using Focal loss $\mathcal{L}_{cls}$ \cite{focalloss}, GIoU loss $\mathcal{L}_{GIoU}$ \cite{giou}, and $\mathcal{L}_{1}$ loss, guiding the self-supervised learning process of the proposed model. The loss $\mathcal{L}$ (i.e., $\mathcal{L}_{u}$ and $\mathcal{L}_{s}$) can be formulated as:
   \begin{equation}
      \mathcal{L} = \mathcal{L}_{cls} + \lambda_{1}\mathcal{L}_{1} + \lambda_{2}\mathcal{L}_{GIoU},
     \label{eq:loss}
   \end{equation}
where $\lambda_{1}$ and $\lambda_{2}$ are the regularization parameters. By adopting this denoising strategy, our approach enhances training stability and improves tracking robustness.

\section{Experiments}

\subsection{Implementation Details}
\label{exp}

   The training data includes LaSOT \cite{lasot}, TNL2K \cite{tnl2k}, and OTB99 \cite{trackNL}.
   The AdamW \cite{adamw} is used to optimize model parameters with initial learning rate of $2.5 \times 10^{-5}$ for the backbone, $2.5 \times 10^{-4}$ for the rest, and set the weight decay to $10^{-4}$.
   The training epochs is set to $150$ epochs. $10$K image pairs are randomly sampled in each epoch.
   The learning rate drops by a factor of $10$ after $120$ epochs.
   The model is conducted on a server with four 80GB A800 GPUs and set the batch size to be $16$.
   We use a ViT-Base \cite{vit} model with DropMAE \cite{DropMAE} pre-trained parameters as the visual encoder.
   The BERT \cite{bert} is used as our language encoder.
   For the input video clip, including three unlabeled frames and one init frame with pseudo-label.
   We present two variants of {\tracker} with different configurations as follows: 1) \textit{{\tracker}-L}: Self-supervised VL tracking with only natural language description. 2) \textit{{\tracker}-V}: Self-supervised visual tracking with only initial bounding box.
   Additionally, our {\tracker}-384 runs at 56 FPS on an A100 GPU.

\begin{table}[t]
    \centering
    \caption{Comparison with state-of-the-art methods on LaSOT${_{\rm{ext}}}$.
    }
    \resizebox{\linewidth}{!}{
    \begin{tabular}{c|l|c|c|ccc}
    \toprule
    \multicolumn{1}{c|}{\multirow{2}{*}{Type}} & \multicolumn{1}{c|}{\multirow{2}{*}{Method}} & \multicolumn{1}{c|}{\multirow{2}{*}{Supervised}} & \multicolumn{1}{c|}{\multirow{2}{*}{Initialize}}
      & \multicolumn{3}{c}{LaSOT${_{\rm{ext}}}$} \\ \cline{5-7}
      & & & & AUC & P${_{\rm{Norm}}}$ & P \\
      \midrule
      \multicolumn{1}{c|}{\multirow{7}{*}{\rotatebox{90}{VL}}} & VLT$_{\rm{TT}}$ \cite{VLT} & Yes & NL+BBox & {48.4} & {59.9} & {54.3} \\ 
      & MMTrack \cite{mmtrack} & Yes & NL+BBox & {49.4} & {59.9} & {55.3} \\
      & All-in-One \cite{Allinone} & Yes & NL+BBox & {54.5} & - & {66.0} \\
      & PJVLT \cite{PJVLT} & Yes & NL+BBox & {48.5} & 58.8 & {55.0} \\
      & MemVLT \cite{MemVLT} & Yes & NL+BBox & {52.1} & {63.3} & {59.8} \\
      \rowcolor{blue!10}
      & \textbf{{\tracker}-L256} & \textit{Init.Lang} & NL & \underline{34.0} & \underline{41.9} & \underline{36.2} \\ 
      \rowcolor{blue!10} 
      & \textbf{{\tracker}-L384} & \textit{Init.Lang} & NL & \textbf{35.2} & \textbf{43.6} & \textbf{37.6} \\
      \midrule
      \multicolumn{1}{c|}{\multirow{7}{*}{\rotatebox{90}{Vision}}} 
      & TransT \cite{transt} & Yes & BBox & 44.8 & - & 52.5 \\
      & STARK \cite{stark} & Yes & BBox & 47.7 & - & 54.9 \\
      & OSTrack \cite{ostrack} & Yes & BBox & 47.4 & 57.3 & 53.3 \\
      & ARTrack \cite{ARTrack} & Yes & BBox & 51.9 & 62.0 & 58.5 \\
      & ODTrack \cite{odtrack} & Yes & BBox & 52.4 & 63.9 & 60.1 \\
      \rowcolor{blue!10}
      & \textbf{\tracker-V256} & \textit{Init.BBox} & BBox & \underline{46.5} & \underline{58.4} & \underline{52.3} \\
      \rowcolor{blue!10}
      & \textbf{\tracker-V384} & \textit{Init.BBox} & BBox & \textbf{49.7} & \textbf{61.9} & \textbf{56.9} \\
    \bottomrule
    \end{tabular}
    }
    \label{tab:lasot_ext}
\end{table}

\subsection{Comparison with the SOTA}

   \textbf{Discussion on VL tracking.}
   As shown in Tab.\ref{tab:results} and Tab.\ref{tab:lasot_ext}, we evaluate the proposed model on four VL tracking benchmarks, including LaSOT \cite{lasot}, LaSOT$_{\rm{ext}}$ \cite{lasot-ext}, TNL2K \cite{tnl2k}, and OTB99 \cite{trackNL}, and compare with existing SOTA trackers. 
   Specifically, under the self-supervised setting, our {\tracker}-L256 demonstrates superior tracking performance compared to ATTracker \cite{ATTracker} initialized with both the initial bounding box and language descriptions, achieving consistent improvements across the TNL2K, LaSOT, and OTB99 datasets. For instance, it outperforms ATTracker by 1.9\%, 3.6\%, and 9.9\% in terms of AUC metrics, respectively.
   Besides, our {\tracker}-L256 achieves competitive performance on LaSOT$_{\rm{ext}}$, with a success score (AUC) of 34.0\%, a normalized precision score (P$_{\rm{Norm}}$) of 41.9\%, and a precision score (P) of 36.2\%.
   These results show that the proposed dynamic token aggregation module helps our {\tracker} to autonomously learn tracking knowledge about target appearance and motion cues, which significantly improves the tracking robustness in complex scenarios.

   \begin{figure}[t]
      \centering
      \includegraphics[width=0.9\linewidth]{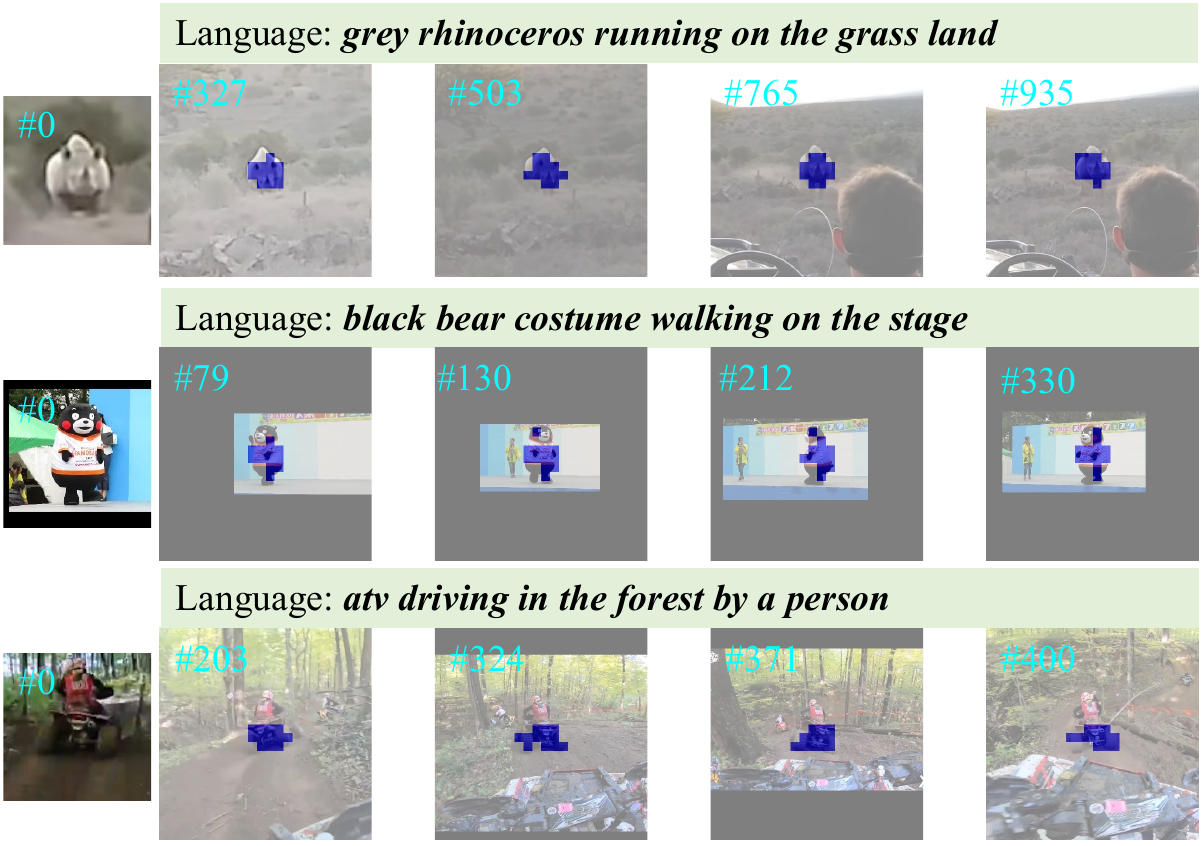}
       \caption{Visualization of the instance tokens sampled from the search frame on LaSOT benchmark.}
       \label{fig:vis_mask}
    \end{figure}

   \textbf{Discussion on visual tracking.}
   Meanwhile, we conducted comprehensive comparisons with numerous SOTA fully-supervised and unsupervised tracking methods across these four VL tracking datasets. As shown in Tab.\ref{tab:results}, our model outperforms all unsupervised tracking algorithms. For instance, on the LaSOT dataset, {\tracker}-V256 surpasses the latest unsupervised tracker, Diff-Tracker, by 16.5\% in terms of AUC score. Furthermore, as shown in Tab.\ref{tab:lasot_ext}, our tracker achieves performance comparable to fully-supervised tracking methods on the LaSOT$_{\rm{ext}}$ dataset. 
   Specifically, {\tracker}-V384 achieves an AUC score of 49.7\%, a P$_{\rm{Norm}}$ score of 61.9\%, and a P score of 56.9\%. 
   These results demonstrate that our self-supervised tracking framework not only delivers superior performance compared to both fully-supervised and unsupervised tracking algorithms but also significantly narrows the performance gap with fully-supervised methods.

\subsection{Ablation Study}

To explore the role of the various components in the proposed framework, we conduct comprehensive ablation studies on the LaSOT \cite{lasot} benchmark.

\textbf{Study on different modules.}
As shown in Sec.\ref{sec:framework}, three critical factors influencing self-supervised tracking performance are the weak-to-strong consistency framework, dynamic token aggregation module (DTA), and denoising training strategy. As shown in Tab.\ref{tab:variant}, removing the DTA (\#2) leads to a 1.1\% drop in AUC score, indicating that dynamically selecting tokens for multi-modal fusion effectively enhances the semantic understanding capability of the model. When the denoising training strategy (\#3) is removed, the AUC slightly decreases by 0.5\%, suggesting that it helps mitigate the interference of noisy samples. Furthermore, removing the weak-to-strong framework (\#1) results in a 0.9\% performance drop, demonstrating that it increases sample diversity and facilitates the self-supervised model in autonomously learning richer representations.

\begin{table}[h]
  \centering
  \caption{Comparison of different modules.}
  \begin{tabular}{c|c|ccc}
    \toprule
    \# & Methods & AUC & P$_{Norm}$ & P \\
    \midrule
    \rowcolor{verylightgray}
    1 & {\tracker} & \textbf{52.5} & \textbf{60.0} & \textbf{52.3} \\
    2 & - weak-to-strong & 51.6 & 59.1 & 51.1 \\
    3 & - DTA & 51.4 & 59.0 & 50.8 \\
    4 & - Denoising & 52.0 & 59.7 & 51.4 \\
    \bottomrule
  \end{tabular}
  \label{tab:variant}
\end{table}

\textbf{Study on the number of search tokens.}
As shown in Tab.\ref{tab:token_num}, we investigate the impact of the number of extracted search tokens on self-supervised tracking performance. Specifically, increasing the number of tokens from 4 to 8 yields a 0.6\% improvement in AUC. However, further increasing the number of tokens leads to a performance drop, indicating that selecting an appropriate number of latent target tokens is crucial. Excessive tokens may introduce noise from non-target regions, thus undermining stability.

\begin{table}[h]
  \centering
  \caption{Comparison of the number of extracted search tokens.}
  \begin{tabular}{c|c|ccc}
    \toprule
    \# & Number of Tokens & AUC & P$_{Norm}$ & P \\
    \midrule
    1 & \textit{4} & 51.9 & 60.0 & 52.0 \\
    \rowcolor{verylightgray}
    2 & \textit{8} & \textbf{52.5} & \textbf{60.0} & \textbf{52.3} \\
    3 & \textit{16} & 52.1 & 59.6 & 51.9 \\
    \bottomrule
  \end{tabular}
  \label{tab:token_num}
\end{table}

\begin{figure}[h]
      \centering
      \includegraphics[width=1\linewidth]{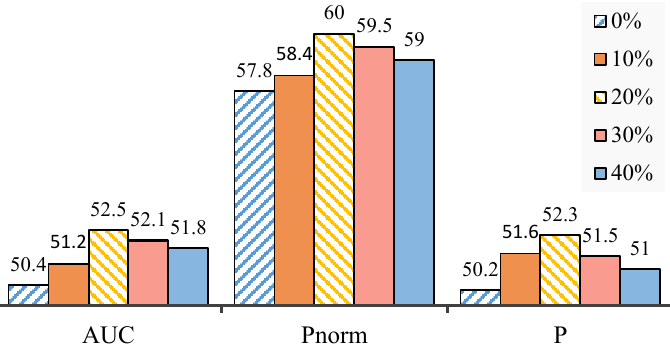}
       \caption{Comparison of different denoising ratios.}
       \label{fig:denoising}
    \end{figure}

\textbf{Study on Denoising Training Strategy.} 
We investigate different noise evaluation methods (i.e., cross-entropy distance and Euclidean distance) and denoising ratios to develop a robust denoising training strategy. As shown in Tab.\ref{tab:distance}, compared to the cross-entropy, Euclidean distance better captures the overall differences between samples and achieves superior tracking performance. Thus, we adopt Euclidean distance, which is more robust to noise, for filtering noisy samples. Furthermore, we conduct comparative experiments by discarding different proportions of noisy samples. As shown in Fig.\ref{fig:denoising}, selecting a 20\% denoising ratio leads to a more robust self-supervised tracker across diverse scenarios. These results indicate that designing an appropriate denoising training strategy is essential for improving the stability of self-supervised learning in noisy environments.

\begin{table}[h]
  \centering
  \caption{Comparison of noise evaluation methods.}
  \begin{tabular}{cc|ccc}
    \toprule
    \# & Evaluation Method & AUC & P${_{\rm{Norm}}}$ & P \\
    \midrule
    1 & \textit{Cross-entropy} & 47.9 & 54.7 & 46.4 \\
    \rowcolor{verylightgray}
    2 & \textit{Euclidean distance} & \textbf{52.5} & \textbf{60.0} & \textbf{52.3} \\
    \bottomrule
  \end{tabular}
  \label{tab:distance}
\end{table}

\textbf{Impact of Pseudo Label Generation.}
We compare different large vision-language models for generating pseudo-label in VL tracking datasets, such as LISA \cite{LISA} and APE \cite{APE}. As shown in Tab.\ref{tab:LVLM}, APE, which incorporates instance-level perception, achieves superior tracking performance compared to LISA, which possesses reasoning segmentation capabilities. This suggests that APE can generate pseudo-bounding boxes that better align with natural language descriptions. These findings highlight the necessity of deploying a robust cross-modal understanding encoder in self-supervised VL tracking framework to enhance the alignment between natural language and visual content.

\begin{table}[h]
  \centering
  \caption{Comparison of different large vision-language models.}
  \begin{tabular}{cc|ccc}
    \toprule
    \# & Vision-Language Model & AUC & P${_{\rm{Norm}}}$ & P \\
    \midrule
    1 & \textit{LISA} & 48.4 & 57.1 & 47.4 \\
    \rowcolor{verylightgray}
    2 & \textit{APE} & \textbf{52.5} & \textbf{60.0} & \textbf{52.3} \\
    \bottomrule
  \end{tabular}
  \label{tab:LVLM}
\end{table}

\subsection{Visualization Analysis}

\begin{figure}[h]
      \centering
      \includegraphics[width=0.9\linewidth]{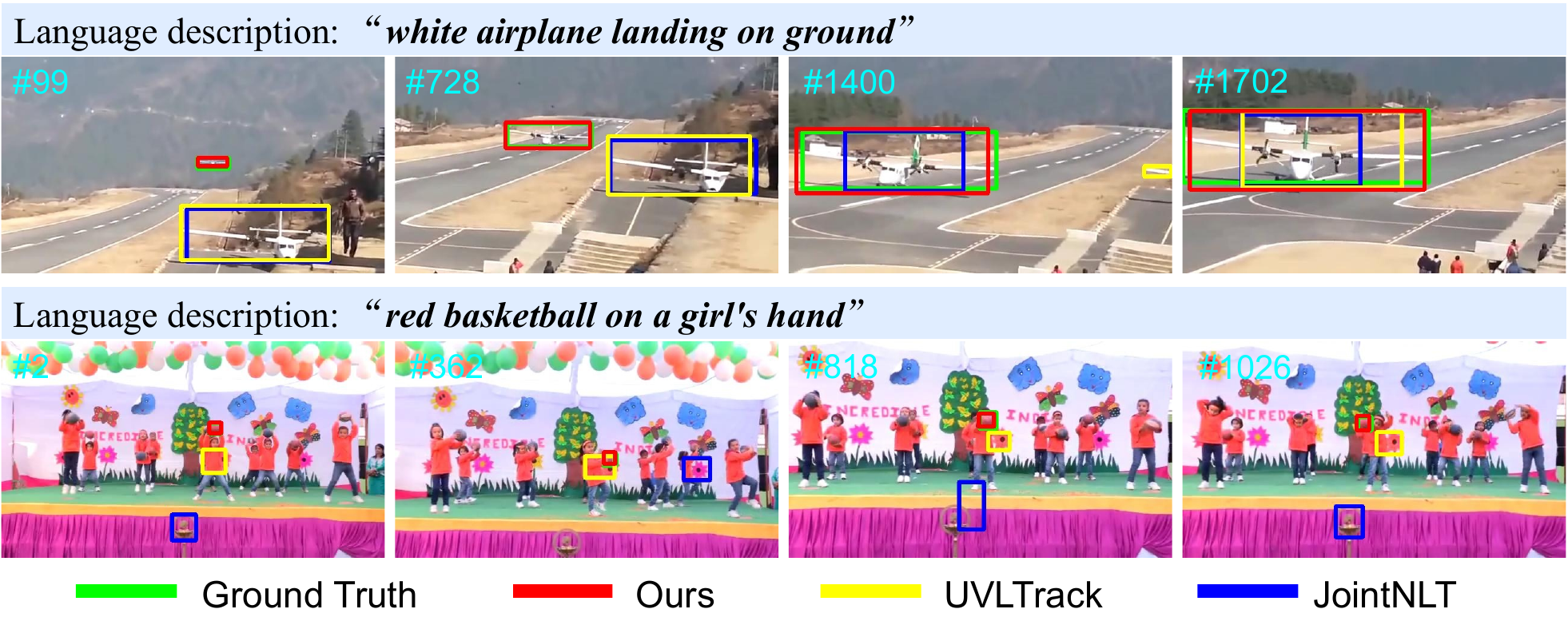}
       \caption{Qualitative comparison results of our tracker on LaSOT.}
       \label{fig:visual}
    \end{figure}

\textbf{Visualization}. As shown in Fig.\ref{fig:vis_mask}, we verify whether the sampled latent tokens from the search frame originate from the target instance. In the third complex scenario, the sampled target tokens are effectively localized on the vehicle, demonstrating that our Dynamic Token Aggregation (DTA) module possesses good target awareness capability. Furthermore, Fig.\ref{fig:visual} presents a visualization comparison of bounding boxes between {\tracker} and the fully supervised VL tracking methods UVLTrack \cite{UVLTrack} and JointNLT \cite{jointNLT}. It can be observed that our tracker exhibits superior performance across various challenging scenarios, even achieving results comparable to those of fully supervised VL trackers.

\section{Conclusion}
\label{conclusion}

We have proposed a self-supervised VL tracking method, {\tracker}, designed to quantify the contribution of natural language descriptions to VL tracking. 
Specifically, within a weak-to-strong consistency self-supervised framework, we have proposed a novel dynamic token aggregation module to efficiently fuse visual and linguistic features.
Besides, a denoising training strategy has been introduced to enhance the stability of self-supervised VL tracking by discarding noisy samples. Extensive experiments have demonstrated the superiority of our approach.

\textbf{Limitation.}
Despite achieving impressive performance, we observe that the generation of pseudo-labels is constrained by the LVLM, which to some extent affects the tracking results. Thus, improving the quality of the acquired target identity information further enhances the performance of self-supervised VL tracking.

\textbf{Acknowledgement.} This work is supported by the Project of Guangxi Science and Technology (No. 2024GXNSFGA010001 and GuiKeFN2504240017), the National Natural Science Foundation of China (No.U23A20383, 62472109 and 62466051), the Guangxi ”Young Bagui Scholar” Teams for Innovation and Research Project, the Research Project of Guangxi Normal University (No. 2025DF001).
{
    \small
    \bibliographystyle{ieeenat_fullname}
    \bibliography{main}
}



\newpage

\section{Appendix}
\textbf{Additional Backgrounds.} 
With the advancement of deep learning techniques \cite{li2025cross,qiu2025convex,zhou2026comem,song1,song2,tcsvt1,peng2024lightweight,pengtowards,lu2023tf,lu2024mace,geng2025survey,zhang2025rfmamba,jin2025masked,dai2025unbiased,yin2025knowledge,he2025reversible,he2025segment} and the potential to eliminate the need for large-scale labeled data, self-supervised tracking has attracted increasing attention from researchers. Taking advantage of intrinsic correlations in unlabeled video data, such as temporal consistency, self-supervised tracking has shown promising results in relatively simple tracking scenarios. However, in long-term complex unlabeled tracking settings, it remains a significant challenge to capture cross-frame motion patterns and to learn robust target representations.

\textbf{Evaluation Metrics.} The tracking performance is evaluated using the toolkit corresponding to the dataset. We follow the evaluation protocol of published datasets and employ three metrics to ensure a fair comparison across various tracking methods, including success score (AUC), normalized precision score (P$_{\rm{Norm}}$), and precision score (P).

\end{document}